%% file: main.tex
\documentclass[sigconf,screen]{acmart}

\usepackage{amsmath}
\usepackage{booktabs} 
\usepackage{algorithmic}
\usepackage{algorithm2e}
\usepackage{wrapfig}
\usepackage{graphicx}
\usepackage{amssymb}
\usepackage{multirow}
\usepackage{subfig}
\usepackage{verbatim}
\usepackage{color}

\setcopyright{rightsretained}

\acmDOI{10.475/123_4}

\acmISBN{123-4567-24-567/08/06}

\acmConference[ADKDD'18]{ADKDD'18}{August 20, 2018}{London, United Kingdom}
\acmYear{2018}
\copyrightyear{2018}

\acmArticle{4}
\acmPrice{15.00}

\begin{document}
\title{Forecasting Granular Audience Size for Online Advertising}

\author{Ritwik Sinha} \affiliation{\institution{Adobe Research}}

\author{Dhruv Singal}
\affiliation{%
 \institution{Adobe Research}
}

 %

\author{Pranav Maneriker}
\affiliation{%
 \institution{Adobe Research}
}

\author{Kushal Chawla}
\affiliation{%
 \institution{Adobe Research}
}

\author{Yash Shrivastava}
\affiliation{%
 \institution{IIT Kharagpur}
}

\author{Deepak Pai}
\affiliation{%
 \institution{Adobe Systems}
}

\author{Atanu R Sinha}
\affiliation{%
 \institution{Adobe Research}
}







\renewcommand{\shortauthors}{Anonymous et al.}
\newcommand\numberthis{\addtocounter{equation}{1}\tag{\theequation}}

\begin{abstract}
 Orchestration of campaigns for online display advertising requires marketers to forecast audience size at the granularity of specific attributes of web traffic, characterized by the categorical nature of all attributes (e.g. \{US, Chrome, Mobile\}). With each attribute taking many values, the very large attribute combination set makes estimating audience size for any specific attribute combination challenging. We modify Eclat, a frequent itemset mining (FIM) algorithm, to accommodate categorical variables. For consequent frequent and infrequent itemsets, we then provide forecasts using time series analysis with  conditional probabilities to aid approximation. An extensive simulation, based on typical characteristics of audience data, is built to stress test our modified-FIM approach. In two real datasets, comparison with baselines including neural network models, shows that our method lowers computation time of FIM for categorical data. On hold out samples we show that the proposed forecasting method outperforms these baselines.


\end{abstract}
%
%
\maketitle

\input{introduction.tex}
\input{related.tex}
\input{overview.tex}
\input{approach.tex}
\input{experiments.tex}

\input{conclusions.tex}
\bibliographystyle{ACM-Reference-Format}
\bibliography{acmart} 

\end{document}

%% file: introduction.tex
\section{Introduction}

The online display advertising (hereafter, display ad) ecosystem has many players that intermediate between publishers and marketers~\cite{Muthukrishnan2009}. For targeting ad campaigns to consumers it is imperative for a marketer to estimate the number of visitors satisfying a set of targeted attribute values in a future time period. Consider one such target-\{Country:US, Browser:Chrome, Device:Mobile\}. The marketer may be interested in predicting the number of advertising bid requests for this target flowing into the Demand Side Platform(DSP) in the following week. This helps optimize spend allocation among various targets in a campaign, as well as helps manage the marketing budget across campaigns.

Forecasting granular-level audience size poses a considerable data mining challenge because of the explosion in the number of possible categorical attribute value combinations. One of our two real world datasets contains $10$ attributes, each taking many values (even $100$ or more), resulting in $\sim10^{20}$ unique targets. While not all combinations are observed in the data, it is still infeasible to store data for all observed combinations and apply time series estimation methods. Notably, forecasting audience size for web traffic is an under-researched area, although programmatic advertising is the subject of growing research, with inroads in diverse topics like bid optimization~\cite{Zhang2014}, targeting~\cite{Goldfarb2011} as well as estimating conversion rate~\cite{Lee2012} and click-through rate~\cite{Zhang2016}. 

In proposing a practicable solution, we develop a three stage approach: first, bringing the problem to a tractable scale using frequent itemset mining (FIM); second, using conditional probability to extend to unobserved targets and third, leveraging time series analysis methods to forecast. Our approach is evaluated on two datasets: first, bid requests data received by a DSP and second, web analytics data of a US publisher. The DSP receives bid requests from multiple Ad Exchanges and serves multiple advertisers. The web analytics data, although from a single publisher, is more feature-rich than the bid requests data. While the two settings are different, the forecasting problem has important commonalities: both datasets comprise historical \emph{time stamped events of consumers} (representing bid requests and page views in respective settings), where each event is defined using values for a set of categorical \emph{attributes}. For each dataset, we forecast the number of events occurring in a given time period for a specific target set defined using values of categorical attributes. Our solution for the first dataset computes and stores the support for $5\times10^5$ frequent itemsets out of a possible $3.84\times10^{18}$ and only about $100$ time series models, and yet, is more accurate than baselines.

Online audience estimation requires forecasts (1) be available for any arbitrary attribute value combination, (2) be frequently updated, and (3) account for temporal variations. Historical time stamped events are used to estimate number of events with specified attribute values in a future time period. Notably, all attributes of web traffic data are categorical and most attributes show long-tailed univariate distributions  (Figure~\ref{fig:univs}). Under this premise, our contributions are: One, we leverage the categorical nature of attributes to efficiently mine frequent attribute combinations from the event database (Section~\ref{sec:approach:fim}) by modifying a leading FIM algorithm to include categorical constraints. This improves performance time and helps meet (2). Two, the mined frequent item (attribute) sets (FIS) are only a small portion of attribute combinations used by firms for targeting. For the non-FIS, which is a very large set, we offer a scalable method for forecasting since the cost of storing all data is prohibitive. Our solution uses an approximation based on conditional probability, storing only on relatively few attribute sets. Three, given a target set definition and a time period in the testing phase, we select an appropriate time series model for predictions and then use information obtained from FIM to obtain estimates of audience size, which meet (3). The approach also estimates for non-FIS, thereby providing estimates for any arbitrary itemset (Section~\ref{sec:approach:estimation}) and satisfies (1). Four, contributing to the FIM literature concerned with categorical variables, we introduce a simulation framework to stress test FIM algorithms. 

%% file: related.tex
\section{Related Work}
The curse of cardinality in web traffic attribute combinations manifests in adverse query time and  massive cost of storing temporal data. Websites need forecasts at granular level of attribute combinations and updated often. While existing FIM algorithms may handle the curse by extracting FIS, that fails to meet the website's needs for forecasts for most other non-frequent itemsets. We bring tools from probability and time series to address these issues. The forecasting problem considered in this work has been explored earlier by Agarwal et al~\cite{agarwal2010forecasting}. However, they use domain knowledge in display advertising to build time series models for a subset of attribute combinations; we use FIM to build a generalizable approach.

The Apriori algorithm~\cite{Agrawal1994} has been extended to Eclat~\cite{Han2000}, FP-Growth~\cite{zaki1997new}, and LCM~\cite{Uno2004} algorithms. The latter three are considered better off-the-shelf algorithms for association rule mining problems~\cite{Borgelt2012}. In further development, ~\cite{Srikant1997,Pei2001} adds category-based constraints to Apriori~\cite{Do2003}. Advancing the work, we add categorical constraints to Eclat and show better performance against other state-of-the-art algorithms.

Time series forecasting is not new ~\cite{Hamilton1994}. Recent attention to search through a class of models to provide forecasts based on best performing models includes Exponential Smoothing~\cite{Hyndman2009}, Automatic ARIMA models~\cite{Hyndman2008} and Prophet~\cite{taylor2017forecasting}. We explore these three and a Neural Net based approach in our experiments.

Our introduction of a new framework for stress testing FIM algorithms draws upon statistical copula~\cite{nelsen1999introduction}, to capture statistical dependencies among categorical variables. Existing data sets for testing FIM algorithms are not built for categorical variables. This approach to the FIM literature is expected to help in testing and comparing suitability of algorithms for data with categorical variables, which are common in web traffic.

%% file: approach.tex
\section{Approach}\label{sec:approach}
Let us define a set of attributes $A =\{A_1, A_2, \dots, A_k\}$, where each $A_l$ takes one of a possible set of values, $V_l$. Let the set of events be $\mathcal{D} = \{ d_1, d_2, \dots, d_n\}$, where each event or \emph{transaction} is defined by value assignments for each attribute, $d_i = \{d_{\mathit{i1}}, d_{\mathit{i2}}, \dots, d_{\mathit{ik}}\}$ where $d_{\mathit{il}} \in V_l$. Additionally, each transaction has a timestamp associated with it. We define a \emph{target definition} as $T = \{t_1, t_2, \dots, t_k\}$, where $t_l \in V_l \cup \{u_l\}$, where $u_l$ is a special marker indicating that $t_l$ can take any value in $V_l$. This marker defines targets where some attributes are left unspecified. A transaction $d_i$ \emph{satisfies} the target definition $T$ if for all $l, d_{\mathit{il}} = t_l$ where $t_l \neq u_l$. The audience estimation problem is formally stated as: given a historic dataset $\mathcal{D}$, estimate the number of events $d_i$ satisfying $T$, in a future time range. 

\subsection{Frequent Itemset Mining}\label{sec:approach:fim}

In frequent itemset mining(FIM), the events could be transactions, as in the case of purchase, or occurrences of audience member on a publisher site, as in our case. The problem is formally stated as follows~\cite{Borgelt2012}. For the set of \emph{transactions} $\mathcal{D}$, such that each transaction is a set of \emph{items}, denote the set of all possible items as $\mathcal{I} = \{\iota_1, \iota_2, \dots,  \iota_m\}$. Hence, each transaction is an \emph{itemset}, $d_p \subseteq \mathcal{I}$. The cover $K(I) \subseteq \mathcal{D}$ of itemset $I \subseteq \mathcal{I}$ is the set of all transactions $\{d_p\} \in \mathcal{D}$ such that $I \subseteq d_p$. The \emph{support} $s(I)$ is the size of $K(I)$, $s(I) = |K(I)|$. The problem is to find all itemsets  $\{I_1, I_2, \dots, I_m\}$ in $\mathcal{D}$ with support more than a threshold $\kappa$. Additional constraints allow more efficient enumeration of frequent itemsets~\cite{Srikant1997}. A constraint is a mapping from the power set of items to a boolean value, $C : 2^{\mathcal{I}} \rightarrow \{\mathit{True}, \mathit{False}\}$. FIM algorithms exploit properties of the support constraint ($s(I) > \kappa$).

A characteristic of online traffic is that the $A_l^\text{th}$ attribute of a transaction $d_p$ takes only one of the values in $V_l$. 
This implies that any itemset which has two or more values for the same attribute must have a zero count, which we encode as the categorical constraint ($CC$). We modify Eclat by checking for $CC$ during the candidate set generation stage. Note that LCM and FP-Growth have both the horizontal and vertical representations of the transactions (explicitly in case of LCM and as the FP-Tree in FP-Growth)~\cite{Borgelt2012}, thus cannot benefit from the inclusion of $CC$. Formally, $\mathit{CC}(I) = \ \mathit{True} \ \text{iff}\  i_l \in V_l \ \forall l$,
where $I$ is the transaction $(i_1, i_2, \dots, i_k)$, and $V_l$ is as defined in Section~\ref{sec:approach}. Constraints can be characterized by some properties such as anti-monotone, succinct, and convertible~\cite{Ng1998}. We state the definitions of two such properties here. 
\begin{definition}{\textbf{Anti Monotone:}}
A constraint $C(\cdot)$ defined on sets is anti-monotone iff for all itemsets $S \subseteq S'$, $C(S) = \mathit{False} \implies C(S') = \mathit{False}$.
\end{definition}
\begin{definition}{\textbf{Succinct:}}
A constraint $C(\cdot)$ defined on sets is succinct iff for all itemsets $I$: $C(I)$ can be expressed as $\forall e \in I$ : $r(e) = \mathit{True}$ for a predicate $r$. 
\end{definition}

$CC$ is anti-monotone and succinct~\cite{Do2003}. Anti-monotone constraints can be applied to a level-wise algorithm, at each level successively~\cite{Do2003}. Moreover, if a constraint is succinct, it is also \emph{pre-counting pushable}. While~\cite{Do2003} applied $CC$ to Apriori, we extend $CC$ to Eclat. This is done by \emph{pre-counting pruning}, that is, $CC$ can be pushed to the stage post the candidate generation phase and prior to support related checks, discarding ineligible candidates. For Eclat, the check is pushed to the stage prior to applying intersections of transaction lists of generated candidates (see Algorithm~\ref{algo:eclatcc}).

\RestyleAlgo{boxruled}

\begin{algorithm}[t]
\DontPrintSemicolon
\SetKwProg{Fn}{Function}{}{}
\tcp{Define $t(I)$: Transaction ID list for itemset $I$}
\tcp{Initial call: $F \gets \varnothing,\ P \gets \left\{ (\{i\}, t(\{i\})) : i \in \mathcal{I},\ |{t(\{i\})}| \geq \kappa \right\}$ }
\Fn{\textsc{EclatCC($P$, $\kappa$, $F$)}}{
    \KwResult{$F$, the set of frequent itemsets}

    \ForAll{$(X_a,t(X_a)) \in P$}{
        \tcp{$X_a$ is a frequent itemset}
        $F \gets F \cup \{(X_a, s(X_a)\}$, $P_a \gets \varnothing$  \\
        \ForAll{$(X_b, t(X_b)) \in P$, with $X_b > X_a$}{
            $X_{ab}$ = $X_a \cup X_b$ \\
            \tcp{Pre-counting pruning}
            \If{$CC(X_{ab})$}{
                $t(X_{ab})\ = \ t(X_{a}) \cap t(X_{b}) $ \\
                \lIf{$s(X_{ab})\ \geq \ \kappa$}{
                    $P_a \gets P_a  \cup \{(X_{ab}, \mathbf{t}(X_{ab})\}$
                }
            }
        }
        \tcp{Recursive call}
        \lIf{$P_a \neq \varnothing$}{\textsc{EclatCC($P_a$, $\kappa$, $F$)}}
    }
}
    \caption{Eclat-CC}
    \label{algo:eclatcc}
\end{algorithm}

\subsection{Audience Estimation}\label{sec:approach:estimation}
The previous section described generation of FIS from $\mathcal{D}_{\tau-l, \tau}$ containing historical transactions in time $(\tau - l, \tau]$. The mined FIS provide $s_{\tau - l, \tau} (T)$, $T$ being a target set satisfying the threshold $\kappa$. The interest lies in the support of $T$ in a future time period $(\tau, \tau + m]$, that is, $s_{\tau, \tau + m}(T)$.
While FIM obtains $s_{\tau - l, \tau}(T)$ for many target sets, forecasting for each requires maintaining highly granular time series data for each, making this infeasible for arbitrary targets, including for non-FIS targets. Our approach requires maintaining a granular time series only for a small number of univariate (single item) targets, and for these targets performing time series forecast that captures seasonal and trend patterns.

 Denote the univariate time series targets as $\mathcal{U}$. Given $\mathcal{D}_{\tau - l, \tau}$, $T$, and a future time period $(\tau, \tau + m]$, we estimate the expected number of events in $(\tau, \tau + m]$. The FIS from $\mathcal{D}_{\tau - l, \tau}$ are stored along with their support. We compute the \emph{best} univariate time series $U$ (see below) to generate predictions for the target $T$, subject to $T \subseteq U$ and $U \in \mathcal{U}$. The predictions for $s_{\tau, \tau + m}(T)$ are as follows:
\begin{equation}
    s_{\tau, \tau + m}(T) = P_{\tau, \tau + m}(T \mid U) \times s_{\tau, \tau + m}(U) \approx P_{\tau - l, \tau}(T \mid U) \times s_{\tau, \tau + m}(U),  \numberthis
    \label{eqn:assump}
\end{equation}
where we use the empirical estimate for $P_{\tau - l, \tau}(T \mid U)$, given by
\begin{align}
\hat{P}_{\tau - l, \tau}(T \mid U) = {\hat{P}_{\tau - l, \tau}(T \cap U)}/{\hat{P}_{\tau - l, \tau}(U)} = {s_{\tau - l, \nonumber \tau}(T)}/s_{\tau - l, \tau}(U), \numberthis
\label{eqn:multiplier}
\end{align}
since $T \subseteq U$. In equation~(\ref{eqn:assump}) we make the assumption that $P_{\tau, \tau + m}(T \mid U) \approx P_{\tau - l, \tau}(T \mid U)$, that is, the conditional probability of $T$ given $U$ remains (almost) constant from the training to the forecasting period. We tested this assumption on the FIS empirically from the two real datasets we work with, and get Pearson correlation $ > 0.99$ between these two quantities for both. 

We approximate $P_{\tau, \tau + m}(T \mid U)$ as $\prod_{i=1}^{k} P_{\tau, \tau + m}(( u_1, \dots,\ t_i, \dots, u_k ) \mid U)$ when $T$ is not frequent, where $u_j$ denotes that the $j^\text{th}$ attribute takes any value in its support.
In other words, we assume conditional independence among the attributes and compute the joint probability as the product of marginal probabilities. 

When $( u_1, \dots,\ t_i,\dots, u_k )$ is frequent, we can use the formulation described in equation~(\ref{eqn:multiplier}). In the other case, we use a threshold probability estimate ${\kappa}/{s_{\tau-l, \tau}(U)}$, where $\kappa$ is the support threshold used for FIM. This is an upper bound on the empirical estimate for this itemset (using equation~(\ref{eqn:multiplier})).

To estimate the second term in equation~(\ref{eqn:assump}), we explore multiple classes of time series models to generate the forecast $\hat{s}_{\tau, \tau+m}(U)$  along with standard deviation for all elements in $\mathcal{U}$ (details in section~\ref{expts:forecasting}). The granularity of forecasts depends on the granularity of the input data. We generate hourly forecasts.

Now, from the set of candidate univariate time series for each target $T$, that is, those which satisfy: (1) $T \subseteq U$, (2) $s_{\tau - l, \tau}(U) \geq \kappa$, we choose the time series with the least error in prediction.
 From this limited set of univariate time series we still generate good predictions, as shown in our experiments. We preselect the univariates at the time of computing the frequent itemsets and choose univariates which satisfy (1) and (2). We choose from possible candidate time series, at prediction time, by minimizing the standard error of the estimate $\sigma \left( \hat{s}_{\tau, \tau + m}(T) \right) = \sigma \left( \hat{P}_{\tau, \tau + m}(T \mid U) \times \hat{s}_{\mathcal{D}_{\tau, \tau + m}}(U) \right) $.

%% file: experiments.tex
\section{Experiments}
\label{sec:expts}

\begin{figure}[t]
    \centering
    \includegraphics[width=1.0\linewidth]{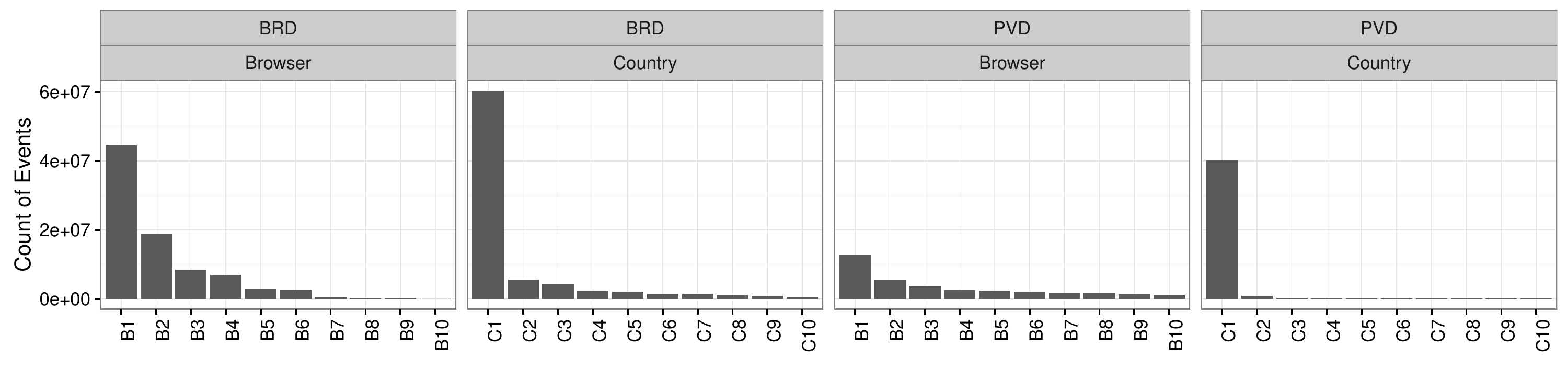}
    \caption{Attribute value frequencies in BRD and PVD}
    \label{fig:univs}
\end{figure}

\textbf{Statistical Copula for a Simulation Framework:}
The current FIM literature offers synthetic datasets~\cite{Agrawal1994} which do not meet the need of emulating categorical nature of web traffic. Our framework fills this gap by creating synthetic data with two important properties: first, the marginal distributions follow structure typically seen in audience data, such as many attributes depicting a long tailed distribution(Figure~\ref{fig:univs}); second, the strong dependence structure common in web traffic be maintained. For example, a type of browser is more likely to be used on a certain operating system. We achieve this by introducing statistical copula~\cite{nelsen1999introduction} into the FIM literature. A copula is a function that joins the multivariate distribution function to their one-dimensional marginals. This approach allows arbitrary marginal distributions while controlling the level of dependence between attributes.

We construct a Gaussian copula from a multivariate normal distribution over $\mathbb{R}^k$, by first specifying a correlation matrix $\mathbf{R}$. We simulate the random vector $\boldsymbol{x} = (x_1, \cdots, x_k)'$ with the multivariate Gaussian cumulative distribution function (CDF) $\Phi_R(\cdot)$ (with correlation matrix $\mathbf{R}$). Then, the vector $(\Phi(x_1), \cdots, \Phi(x_k))'$ (where $\Phi(\cdot)$ is the univariate normal CDF) has marginal distributions which are uniform in $[0,1]$ and a Gaussian copula which captures the dependence. Finally, to achieve the target distributions $F_i(\cdot)$, we perform the transformation $\boldsymbol{y} = (F^{-1}_1(\Phi(x_1)), \cdots, F^{-1}_k(\Phi(x_k)))'$, where $F^{-1}_i(\cdot)$ is the inverse CDF corresponding to $F_i(\cdot)$. The resulting vector $\boldsymbol{y}$ has the desired marginals with a given dependence structure.  

We are still left with deciding two quantities, $\mathbf{R}$ and $F_i(\cdot)$. Experiments with long tailed distributions show a good way to select $F_i(\cdot)$ - base it on the observed multinomial distribution of attribute values. We base the marginals on typically observed distributions in real data (Figure~\ref{fig:univs}). To choose $\mathbf{R}$, we make use of the structure of the observed data. We ensure that the association matrices for the real and simulated data follow a similar pattern. 


With the goal of testing the robustness of FIM approaches and for comparison among them, we vary the following parameters in the synthetic data. First, the number of attributes ($k$) is $8$, $16$ or $32$. Next, the association is either as observed in real data or the off-diagonal elements are half of their values (these are referred to as `high' or `low' correlation). Next, we modify the multinomial marginal distributions to either be long-tailed (`steep') or uniform (`flat').  

\begin{table}[t]

    \caption{Distinct values for attributes in the real datasets}
    \centering
\scalebox{0.8}{    
\begin{tabular}{lr@{\hskip 0.1in}lr}
\toprule
\multicolumn{2}{c}{BRD}   & \multicolumn{2}{c}{PVD}  \\
Attribute & Unique Values & Attribute & Unique Values \\
\midrule
\texttt{ad\_exchange}		&	8		&	\texttt{browser}	&	876\\
\texttt{browser}			&	100		&	\texttt{color\_depth}	&	8\\
\texttt{country}			&	233		&	\texttt{country}	&	233\\
\texttt{device\_family}		&	23141	&	\texttt{domain}	&	61684\\
\texttt{device}				&	3		&	\texttt{language}	&	153\\
\texttt{os}					&	55		&	\texttt{os}	&	257\\
\texttt{region}				&	2598	&	\texttt{ref\_type}	&	7\\
\texttt{slot\_size}			&	693		&	\texttt{region}	&	1043\\
\texttt{slot\_visibility}	&	3		&	\texttt{resolution}	&	448\\
&& \texttt{visit\_number}	&	12884\\
\bottomrule
\end{tabular}
}
\label{tab:distinct}
\end{table}

\textbf{Bid Request Dataset (BRD):}
This dataset arises in the ecosystem where the publisher seeks competitive bids using Ad Exchanges and a Real-Time Bidding (RTB) platform. The publisher delivers the consumer's information, comprising attributes, for real time bidding by marketers seeking consumers matching those attributes. The training data comprise logs from $26$ March to $31$ March $2017$, and the testing data comprise logs for $1$ April, 2017. Around $97$ million bid request events are present, large enough for valid experiments. We have $86$ million and $11$ million bid requests in the training and the testing periods respectively. 
Each event has $9$ attributes (Table~\ref{tab:distinct}), a time stamp, 
and most attributes have a substantial number of distinct values. The number of possible attribute combinations is $3.84\times10^{18}$. The histogram for two attributes is presented in  Figure~\ref{fig:univs}. A similar long tailed distribution exists across all attributes.

\textbf{Page View Dataset (PVD):}
This dataset comes from a publisher, where the publisher sells the consumer's information directly to marketers based on contractual pricing~\cite{Roels2009}. For each page view, the publisher matches the consumer's attributes to those desired by marketers and then offers it to a matched marketer. The contractual mechanism is less studied. Our work applies to both competitive bidding and contractual pricing. This second dataset affords generalization of our approach. The training data comprise $48$ million page views from $31$ March to $5$ April 2017, and the testing data comprise $8$ million for $6$ April. We refer to this dataset as PVD. The dataset has $10$ attributes, some with a large number of distinct values (Table~\ref{tab:distinct}), leading to a total $1.67\times10^{26}$ possible itemsets. As in BRD, attributes display a long-tailed distribution (Figure~\ref{fig:univs}).  

\begin{figure}[t]
\centering
\includegraphics[width=1.0\linewidth]{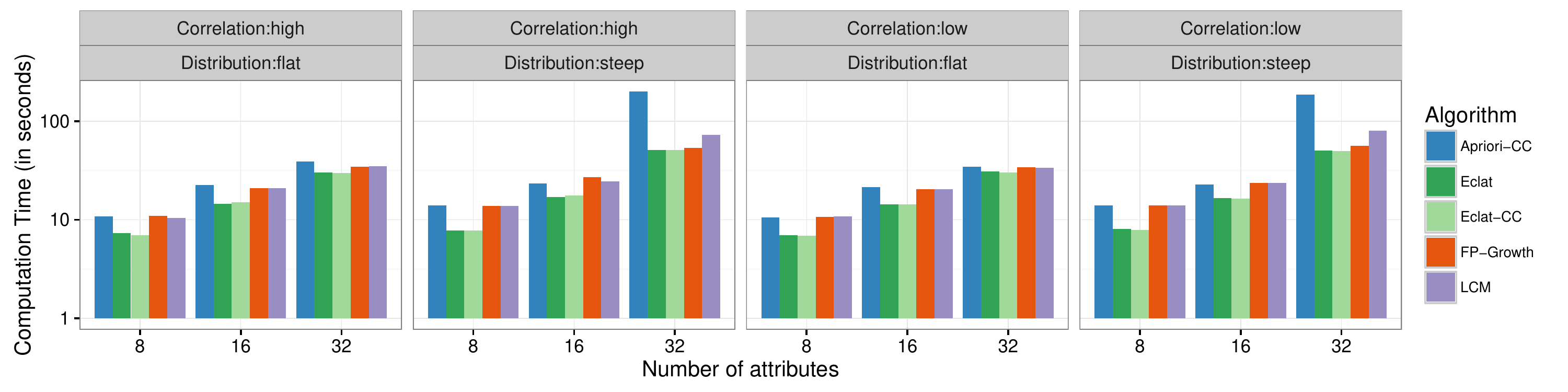}
\caption{Computation time of FIM algorithms on synthetic data. Average time from three runs, presented for different number of attributes, correlation across attributes and marginal  distributions, for support of 10\% (other support levels not displayed in the interest of space).}
\label{fig:sim:results}
\end{figure}

\subsection{Frequent Itemset Mining}\label{subsec:expts-fim}
We perform experiments on the synthetic data and two real datasets. The experiments are carried out on a machine with 16GB RAM and 3.5GHz CPU running a Linux distribution. The algorithms included in our analysis are Apriori-CC~\cite{Do2003}, Eclat~\cite{zaki1997new}, Eclat-CC, LCM~\cite{Uno2004} and FP-Growth~\cite{Han2000}. We follow or extend the implementation of Borgelt~\cite{Borgelt2012} for these algorithms and record the computation time averaged over $3$ runs.

The methods are first compared on the synthetic data (Figure~\ref{fig:sim:results}). We present results for two levels of correlation and two univariate distribution patterns, across three different number of attributes. For each combination, $10$ million events are generated. We make a few broad observations. First, as expected, a higher number of attributes makes the problem more challenging, as reflected in increased computation times. Second, lower correlation leads to a limited decrease in the computation times. Third, having a steep distribution in the univariates leads to higher running times than having flat (equally likely) marginals. This happens because steep distribution and higher correlation lead to higher number of itemsets meeting the threshold, and hence leading to longer run times.

\begin{table*}
    \centering
     \caption{Comparison of FIM algorithms. The average computation time (in seconds) from three runs of the algorithm.}
        \begin{tabular}{l l r r r r r}
        \toprule
        \textbf{Dataset}     & \textbf{Support} & \textbf{Apriori-CC} & \textbf{Eclat} & \textbf{Eclat-CC} & \textbf{FP-Growth} & \textbf{LCM} \\ \hline
        \multirow{3}{*}{PVD} & 1\%              & \textbf{70.7}                & 83.2               & 76.8                  & 71.7               & 72.5         \\
                             & 5\%              & 75.5                & 46.6               & \textbf{46.0}         & 66.4               & 66.2         \\
                             & 10\%             & 71.8                & \textbf{41.9}      & 42.1                  & 59.9               & 59.1         \\ \hline
        \multirow{3}{*}{BRD} & 1\%              & 160.1               & 112.0              & \textbf{105.4}        & 165.1              & 167.5        \\
                             & 5\%              & 167.9               & 80.1               & \textbf{78.7}         & 154.3              & 151.0        \\
                             & 10\%             & 155.6               & 73.7               & \textbf{73.0}         & 135.8              & 137.6 \\
        \bottomrule
        \end{tabular}
    \label{tab:fim}
\end{table*}

In comparing the algorithms, some of the findings are: one, Eclat-CC performs better than unconstrained Eclat, on average; which itself performs better than Apriori-CC. Considering average ranks across different scenarios, the performance of algorithms in decreasing order is -- Eclat-CC, Eclat, LCM, FP-Growth, Apriori-CC. Thus, incorporating CC into Eclat, leads to an algorithm that performs better than the other state-of-the-art algorithms.

On the real data BRD, we find that (Table~\ref{tab:fim}) Eclat-CC is between $2\%$ and $7\%$ better than the next best algorithm, and between $30\%$ and $50\%$ faster than FP-Growth and LCM. On the other real data PVD, Eclat-CC is the best algorithm on a support of $5\%$, while being close to the best algorithm (Eclat) on a support of $10\%$. Moreover, in the case of low support ($1\%$), Eclat-CC performs somewhere in between the best algorithm (Apriori-CC) and Eclat. Thus, using categorical constraints into FIM algorithms leads to more efficient implementations in audience size estimation. It is worth noting that the training data for BRD contains $86$ million events, larger than the other datasets analyzed, suggesting that the gains for incorporating CC may be more pronounced for larger datasets. We test this hypothesis with a simulated dataset of $200$ million events and $8$ attributes, and find that Eclat-CC is $9\%$ better than Eclat and $25\%$ better than FP-Growth. 
\subsection{Audience Forecasting}\label{expts:forecasting}
To evaluate the accuracy of forecasts, we compare our approach with a na\"ive, but feasible baseline (FB), an accurate, but also an infeasible baseline using individual target time series (TS) and a machine learning based method. This comparison across both BRD and PVD datasets, is done on two different target sets - FIS and IFIS(Infrequent-FIS). For \textit{FIS}, the support or threshold value is set at $0.01\%$ of the dataset size throughout. We find $0.5$ million and $0.7$ million FIS in BRD and PVD, respectively. We sample $500$ FIS from each dataset with a probability proportional to the support of the itemset, ensuring that itemsets of varying supports are included in the sample.
For \textit{IFIS}, we sample $500$ infrequent itemsets, among those with less than $0.01\%$ support. We now describe our baseline approaches. 

\textbf{Individual Target Time Series based infeasible baseline (TS)}: Entire time series is stored for all $500$ itemsets in FIS and in IFIS. Forecasts are generated directly by modeling the time series for each itemset, without using conditional probabilities and univariate. This baseline is not bounded by computation time or storage requirements for time series for millions of itemsets. We use it as a boundary condition baseline to compare our approach. 

\textbf{Feasible Baseline (FB)}: We find all univariate itemsets satisfying a given threshold (of $0.5\%$). For each of these, we obtain the hourly counts as a percentage of the global counts for that hour. For such time series, we train a model, so that we can forecast the fraction of hourly global count represented by the respective univariate itemset. We also maintain the global time series, for target $G=(u_1, \cdots, u_k)$, where $u_l$ denotes the $l^\text{th}$ attribute taking any value. For a target $T = (t_1, t_2, ..., t_k)$ we predict the hourly count estimate as $\{P((t_1, u_2, \ldots, u_k)) \times \ldots \times P((u_1, u_2, \ldots, t_k)) \times \hat{s}(G)\}$. Thus, the estimate is obtained by multiplying the global time series forecast by the estimates of percentages of each univariate, obtained from the time series. For univariate values where we do not have a time series, we assume that the percentage varies up to the threshold value used (which in our case corresponds to the interval $[0, 0.005]$). This gives us a ranged estimate, which we average to get the point estimate.

\input{ml-baseline.tex}

We generate time series forecasts for univariate targets in $\mathcal{U}$ (from Section~\ref{sec:approach:estimation}) using four methods -- Exponential Smoothing (ETS), Automatic ARIMA (ARIMA), Neural Network Autoregression (NNAR) and Prophet. We use the respective R packages to automatically choose the best hyper parameters for our time series methods. We use $6$ days of hourly data to train and offer hourly forecasts for the seventh day, capturing daily seasonality. The methods are evaluated using average Mean Absolute Percentage Error (MAPE). Based on superior performance of ETS (Table~\ref{tab:timeseries}), we decide to use it as the time series model for all evaluations.

Figure~\ref{fig:mapes} shows the results. In box plots, bars of the same color denote results on the same target sets, by data set. Mean and median of MAPE across all itemsets are denoted by dashed and solid horizontal lines, respectively. Mean MAPEs for FIS in BRD and PVD, $29\%$ and $25\%$, are lower than Mean MAPEs for FIS-TS in both data, although not for medians, reflecting higher variability of FIS-TS (higher spread in box plot). Hence, we claim that the proposed approach is better in terms of mean MAPEs, than the infeasible baseline. Similarly, the proposed approach always performs better than the feasible, but na\"ive baseline (FB), for both FIS and IFIS; the effect being stronger for FIS. The bad performance of IFIS-TS for PVD may be due to fewer data points of page views for infrequent itemsets. The higher MAPEs for IFIS vs. FIS is due to IFIS itemsets having at most $31$ events every hour on average, a small sample to obtain good estimates. Surprisingly, even in small itemsets, our approach that assumes conditional independence, compares reasonably with IFIS-TS. 

\begin{table}
    \centering
    \begin{minipage}{0.8\linewidth}
        \centering
            \caption{MAPEs for univariate time series\protect\footnote{We also explored a Long Short-Term Memory (LSTM) based time series model, but this failed to provide acceptable accuracy.}}
            \centering
            \begin{tabular}{l l l l l l}
                \toprule
                \textbf{Method} & \textbf{BRD} &  \textbf{PVD} & \textbf{Method} & \textbf{BRD} &  \textbf{PVD} \\
                \midrule
                ETS & 23.2 & 13.6 & NNAR & 24.4 &  17.2 \\
                ARIMA & 32.2 & 17.6 & Prophet & 23.6 & 26.5 \\
                \bottomrule
            \end{tabular}
            \label{tab:timeseries}
        \centering
    \end{minipage}
    \begin{minipage}{0.45\textwidth}
        \centering
        \includegraphics[width=0.8\linewidth]{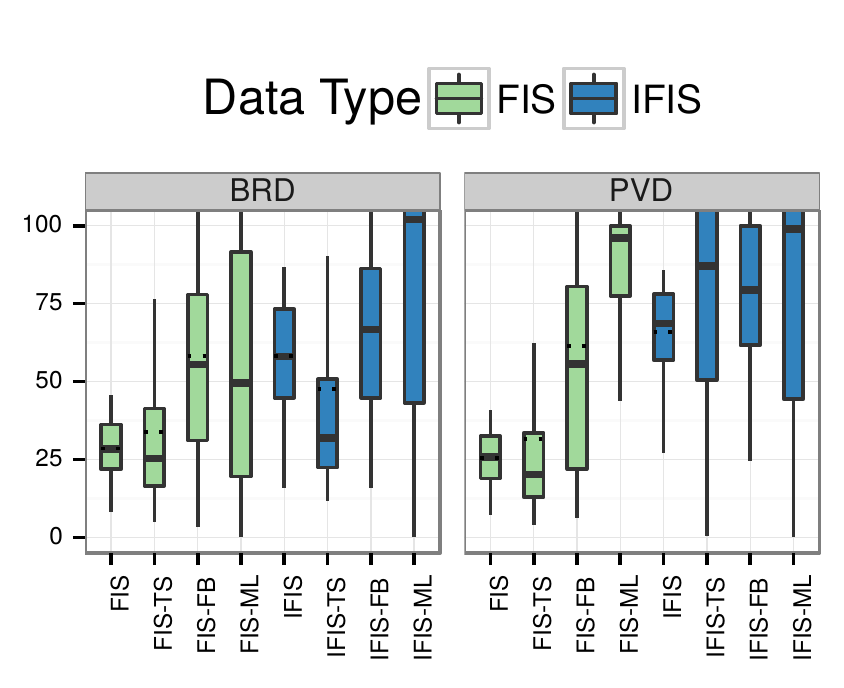}
        \captionof{figure}{MAPEs (Y-axis) for forecasting: Solid and dashed horizontal lines are median and mean. Comparison is relevant across boxes of same color.}
        \label{fig:mapes}
    \end{minipage}
\end{table}
MAPEs are benchmarked against~\cite{hyndman2002state} where ETS produces MAPEs between 10 and 20\% for time series in M3 competition~\cite{makridakis2000m3}. Our MAPEs for univariate time series targets, tasks comparable to the competition, are 14 to 23\%. The audience estimation task is more challenging since forecasts are for thousands of attribute combinations, without recording the time series for each. Our MAPE values under $30\%$ is likely to be acceptable in practice.

%% file: ml-baseline.tex
\textbf{Machine Learning Baseline (ML)}: We modify the datasets to remodel audience forecasting task as a supervised learning problem. To achieve this, we create a training set by sampling 5,000 itemsets from FIS mined at 0.01\% from both PVD and BRD, by sampling with probability proportional to the support, ensuring that itemsets of varying supports are included in the sample (similar to FIS target sets). We collect hourly counts for these itemsets, throughout the training and testing periods. Each row of each data set consists of the itemset, hour of the day and a count of transactions (page views/bid requests) satisfying the itemset in that hour. We drop the day of the week attribute, since our data set is limited to a single week and capturing weekly seasonality is not possible in such a situation. Following the construction of this derived dataset, the forecasting problem is reduced to a regression problem, with a categorical input (itemset, hour) and the output being count of transactions. However, since the total number of levels across various attributes is large (ranging into a few thousands), it is intractable for machine learning models to capture interactions among attributes. Hence, we group all attribute values for which we do not have univariate time series, i.e. present in less than 0.5\% of the dataset, into a new level.

The model is first trained on a subset of sampled FIS, and the trained model is used to make predictions for the same benchmark set as other baselines. The model is a multi-layer fully connected network, with dropout, and with an additional embedding layer in the input, implemented in PyTorch~\cite{paszke2017automatic}. Categorical inputs are mapped to columns of the embedding layer, and then passed through the network to make predictions to minimize MAPE. Hyperparameters are chosen optimally using \texttt{hyperopt}\footnote{https://github.com/hyperopt/hyperopt} library, by considering hyperparameter space spanning embedding layer dimension $ \in  \{32, 64, \dots, 128\}$, dropout $\in [0.0, 0.8]$, and number of layers $\in \{1, 2, 3, 4\}$.  
%

Each parameter is sampled uniformly from the corresponding parameter space. We use 6 days of data to train each model, and optimize the hyperparameters according to the MAPE for the $7^{th}$ day using the \texttt{TPE} algorithm for guiding the search over the hyperparameter space across 1000 trials, with 10 epochs per trial. This search leads to a 3 layer model, with layer dimensions 384, 192, and 64,  a dropout of 0.05 and an embedding dimension of 128. With this model, we generate predictions for the ML baseline, and the results are shown in Figure~\ref{fig:mapes}. We see that the model obtained by this process performs worse than our approach across all experiments. 

%% file: conclusions.tex
\section{Conclusion}

Knowing the likely size of audience segments for web traffic can help websites better plan their ad campaign. Audience forecasting is challenging because of the combinatorial explosion in attribute values, each of which could be a relevant target audience. We address this problem with a combination of frequent itemset mining and time series modeling. We are able to achieve good accuracy levels on real datasets from two use cases within online display ad and compare our results with three baseline approaches. We also give a novel FIM approach, specific to categorical characteristics of audience data. We demonstrate the superior performance of this approach over state of the art algorithms by proposing a new simulation framework.


%% file: main.bbl

\begin{thebibliography}{00}


\ifx \showCODEN    \undefined \def \showCODEN     #1{\unskip}     \fi
\ifx \showDOI      \undefined \def \showDOI       #1{#1}\fi
\ifx \showISBNx    \undefined \def \showISBNx     #1{\unskip}     \fi
\ifx \showISBNxiii \undefined \def \showISBNxiii  #1{\unskip}     \fi
\ifx \showISSN     \undefined \def \showISSN      #1{\unskip}     \fi
\ifx \showLCCN     \undefined \def \showLCCN      #1{\unskip}     \fi
\ifx \shownote     \undefined \def \shownote      #1{#1}          \fi
\ifx \showarticletitle \undefined \def \showarticletitle #1{#1}   \fi
\ifx \showURL      \undefined \def \showURL       {\relax}        \fi
\providecommand\bibfield[2]{#2}
\providecommand\bibinfo[2]{#2}
\providecommand\natexlab[1]{#1}
\providecommand\showeprint[2][]{arXiv:#2}

\bibitem[\protect\citeauthoryear{Agarwal, Chen, Lin, Shanmugasundaram, and
  Vee}{Agarwal et~al\mbox{.}}{2010}]%
        {agarwal2010forecasting}
\bibfield{author}{\bibinfo{person}{Deepak Agarwal}, \bibinfo{person}{Datong
  Chen}, \bibinfo{person}{Long-ji Lin}, \bibinfo{person}{Jayavel
  Shanmugasundaram}, {and} \bibinfo{person}{Erik Vee}.}
  \bibinfo{year}{2010}\natexlab{}.
\newblock \showarticletitle{Forecasting high-dimensional data}. In
  \bibinfo{booktitle}{{\em ACM SIGMOD 2010}}.
\newblock


\bibitem[\protect\citeauthoryear{Agrawal and Srikant}{Agrawal and
  Srikant}{1994}]%
        {Agrawal1994}
\bibfield{author}{\bibinfo{person}{Rakesh Agrawal} {and}
  \bibinfo{person}{Ramakrishnan Srikant}.} \bibinfo{year}{1994}\natexlab{}.
\newblock \showarticletitle{Fast Algorithms for Mining Association Rules in
  Large Databases}. In \bibinfo{booktitle}{{\em VLDB '94}}.
\newblock


\bibitem[\protect\citeauthoryear{Borgelt}{Borgelt}{2012}]%
        {Borgelt2012}
\bibfield{author}{\bibinfo{person}{Christian Borgelt}.}
  \bibinfo{year}{2012}\natexlab{}.
\newblock \showarticletitle{Frequent item set mining}.
\newblock \bibinfo{journal}{{\em Wiley Interdisciplinary Reviews: Data Mining
  and Knowledge Discovery\/}} \bibinfo{volume}{2}, \bibinfo{number}{6}
  (\bibinfo{year}{2012}).
\newblock


\bibitem[\protect\citeauthoryear{Do, Hui, and Fong}{Do et~al\mbox{.}}{2003}]%
        {Do2003}
\bibfield{author}{\bibinfo{person}{Tien~Dung Do}, \bibinfo{person}{Siu~Cheung
  Hui}, {and} \bibinfo{person}{Alvis Fong}.} \bibinfo{year}{2003}\natexlab{}.
\newblock \showarticletitle{Mining Frequent Itemsets with Category-Based
  Constraints}. In \bibinfo{booktitle}{{\em Discovery Science: International
  Conference}}.
\newblock


\bibitem[\protect\citeauthoryear{Goldfarb and Tucker}{Goldfarb and
  Tucker}{2011}]%
        {Goldfarb2011}
\bibfield{author}{\bibinfo{person}{Avi Goldfarb} {and}
  \bibinfo{person}{Catherine Tucker}.} \bibinfo{year}{2011}\natexlab{}.
\newblock \showarticletitle{Online Display Advertising: Targeting and
  Obtrusiveness}.
\newblock \bibinfo{journal}{{\em Marketing Science\/}} \bibinfo{volume}{30},
  \bibinfo{number}{3} (\bibinfo{year}{2011}), \bibinfo{pages}{389--404}.
\newblock


\bibitem[\protect\citeauthoryear{Hamilton}{Hamilton}{1994}]%
        {Hamilton1994}
\bibfield{author}{\bibinfo{person}{J.D. Hamilton}.}
  \bibinfo{year}{1994}\natexlab{}.
\newblock \bibinfo{booktitle}{{\em Time Series Analysis}}.
\newblock \bibinfo{publisher}{Princeton University Press}.
\newblock
\showISBNx{9780691042893}
\showLCCN{lc93004958}


\bibitem[\protect\citeauthoryear{Han, Pei, and Yin}{Han et~al\mbox{.}}{2000}]%
        {Han2000}
\bibfield{author}{\bibinfo{person}{Jiawei Han}, \bibinfo{person}{Jian Pei},
  {and} \bibinfo{person}{Yiwen Yin}.} \bibinfo{year}{2000}\natexlab{}.
\newblock \showarticletitle{Mining frequent patterns without candidate
  generation}. In \bibinfo{booktitle}{{\em ACM SIGMOD}}.
\newblock


\bibitem[\protect\citeauthoryear{Hyndman and Khandakar}{Hyndman and
  Khandakar}{2008}]%
        {Hyndman2008}
\bibfield{author}{\bibinfo{person}{Rob Hyndman} {and} \bibinfo{person}{Yeasmin
  Khandakar}.} \bibinfo{year}{2008}\natexlab{}.
\newblock \showarticletitle{Automatic Time Series Forecasting: The forecast
  Package for R}.
\newblock \bibinfo{journal}{{\em Journal of Statistical Software, Articles\/}}
  \bibinfo{volume}{27}, \bibinfo{number}{3} (\bibinfo{year}{2008}).
\newblock


\bibitem[\protect\citeauthoryear{Hyndman, Koehler, Ord, and Snyder}{Hyndman
  et~al\mbox{.}}{2008}]%
        {Hyndman2009}
\bibfield{author}{\bibinfo{person}{Rob Hyndman}, \bibinfo{person}{Anne~B
  Koehler}, \bibinfo{person}{J~Keith Ord}, {and} \bibinfo{person}{Ralph~D
  Snyder}.} \bibinfo{year}{2008}\natexlab{}.
\newblock \bibinfo{booktitle}{{\em Forecasting with exponential smoothing: The
  state space approach}}.
\newblock \bibinfo{publisher}{Springer}.
\newblock


\bibitem[\protect\citeauthoryear{Hyndman, Koehler, Snyder, and Grose}{Hyndman
  et~al\mbox{.}}{2002}]%
        {hyndman2002state}
\bibfield{author}{\bibinfo{person}{Rob~J Hyndman}, \bibinfo{person}{Anne~B
  Koehler}, \bibinfo{person}{Ralph~D Snyder}, {and} \bibinfo{person}{Simone
  Grose}.} \bibinfo{year}{2002}\natexlab{}.
\newblock \showarticletitle{A state space framework for automatic forecasting
  using exponential smoothing methods}.
\newblock \bibinfo{journal}{{\em International Journal of Forecasting\/}}
  \bibinfo{volume}{18}, \bibinfo{number}{3} (\bibinfo{year}{2002}),
  \bibinfo{pages}{439--454}.
\newblock


\bibitem[\protect\citeauthoryear{Lee, Orten, Dasdan, and Li}{Lee
  et~al\mbox{.}}{2012}]%
        {Lee2012}
\bibfield{author}{\bibinfo{person}{Kuang-chih Lee}, \bibinfo{person}{Burkay
  Orten}, \bibinfo{person}{Ali Dasdan}, {and} \bibinfo{person}{Wentong Li}.}
  \bibinfo{year}{2012}\natexlab{}.
\newblock \showarticletitle{Estimating Conversion Rate in Display Advertising
  from Past Performance Data}. In \bibinfo{booktitle}{{\em ACM SIGKDD
  International Conference on Knowledge Discovery and Data Mining}}.
\newblock


\bibitem[\protect\citeauthoryear{Makridakis and Hibon}{Makridakis and
  Hibon}{2000}]%
        {makridakis2000m3}
\bibfield{author}{\bibinfo{person}{Spyros Makridakis} {and}
  \bibinfo{person}{Michele Hibon}.} \bibinfo{year}{2000}\natexlab{}.
\newblock \showarticletitle{The M3-Competition: results, conclusions and
  implications}.
\newblock \bibinfo{journal}{{\em International journal of forecasting\/}}
  \bibinfo{volume}{16}, \bibinfo{number}{4} (\bibinfo{year}{2000}).
\newblock


\bibitem[\protect\citeauthoryear{Muthukrishnan}{Muthukrishnan}{2009}]%
        {Muthukrishnan2009}
\bibfield{author}{\bibinfo{person}{S. Muthukrishnan}.}
  \bibinfo{year}{2009}\natexlab{}.
\newblock \showarticletitle{Ad Exchanges: Research Issues}. In
  \bibinfo{booktitle}{{\em Proceedings of the 5th International Workshop on
  Internet and Network Economics}} {\em (\bibinfo{series}{WINE '09})}.
\newblock


\bibitem[\protect\citeauthoryear{Nelsen}{Nelsen}{1999}]%
        {nelsen1999introduction}
\bibfield{author}{\bibinfo{person}{Roger~B Nelsen}.}
  \bibinfo{year}{1999}\natexlab{}.
\newblock \showarticletitle{Introduction}.
\newblock In \bibinfo{booktitle}{{\em An Introduction to Copulas}}.
  \bibinfo{publisher}{Springer}, \bibinfo{pages}{1--4}.
\newblock


\bibitem[\protect\citeauthoryear{Ng, Lakshmanan, Han, and Pang}{Ng
  et~al\mbox{.}}{1998}]%
        {Ng1998}
\bibfield{author}{\bibinfo{person}{Raymond~T. Ng}, \bibinfo{person}{Laks V.~S.
  Lakshmanan}, \bibinfo{person}{Jiawei Han}, {and} \bibinfo{person}{Alex
  Pang}.} \bibinfo{year}{1998}\natexlab{}.
\newblock \showarticletitle{Exploratory Mining and Pruning Optimizations of
  Constrained Associations Rules}. In \bibinfo{booktitle}{{\em In the 1998 ACM
  SIGMOD International Conference on Management of Data}}.
\newblock


\bibitem[\protect\citeauthoryear{Paszke, Gross, Chintala, Chanan, Yang, DeVito,
  Lin, Desmaison, Antiga, and Lerer}{Paszke et~al\mbox{.}}{2017}]%
        {paszke2017automatic}
\bibfield{author}{\bibinfo{person}{Adam Paszke}, \bibinfo{person}{Sam Gross},
  \bibinfo{person}{Soumith Chintala}, \bibinfo{person}{Gregory Chanan},
  \bibinfo{person}{Edward Yang}, \bibinfo{person}{Zachary DeVito},
  \bibinfo{person}{Zeming Lin}, \bibinfo{person}{Alban Desmaison},
  \bibinfo{person}{Luca Antiga}, {and} \bibinfo{person}{Adam Lerer}.}
  \bibinfo{year}{2017}\natexlab{}.
\newblock \showarticletitle{Automatic differentiation in PyTorch}. In
  \bibinfo{booktitle}{{\em NIPS-W}}.
\newblock


\bibitem[\protect\citeauthoryear{Pei, Han, and Lakshmanan}{Pei
  et~al\mbox{.}}{2001}]%
        {Pei2001}
\bibfield{author}{\bibinfo{person}{Jian Pei}, \bibinfo{person}{Jiawei Han},
  {and} \bibinfo{person}{L.~V.~S. Lakshmanan}.}
  \bibinfo{year}{2001}\natexlab{}.
\newblock \showarticletitle{Mining frequent itemsets with convertible
  constraints}. In \bibinfo{booktitle}{{\em International Conference on Data
  Engineering}}.
\newblock


\bibitem[\protect\citeauthoryear{Roels and Fridgeirsdottir}{Roels and
  Fridgeirsdottir}{2009}]%
        {Roels2009}
\bibfield{author}{\bibinfo{person}{Guillaume Roels} {and}
  \bibinfo{person}{Kristin Fridgeirsdottir}.} \bibinfo{year}{2009}\natexlab{}.
\newblock \showarticletitle{Dynamic revenue management for online display
  advertising}.
\newblock \bibinfo{journal}{{\em Journal of Revenue and Pricing Management\/}}
  (\bibinfo{year}{2009}).
\newblock


\bibitem[\protect\citeauthoryear{Srikant, Vu, and Agrawal}{Srikant
  et~al\mbox{.}}{1997}]%
        {Srikant1997}
\bibfield{author}{\bibinfo{person}{Ramakrishnan Srikant}, \bibinfo{person}{Quoc
  Vu}, {and} \bibinfo{person}{Rakesh Agrawal}.}
  \bibinfo{year}{1997}\natexlab{}.
\newblock \showarticletitle{Mining Association Rules with Item Constraints}. In
  \bibinfo{booktitle}{{\em KDD'97}}.
\newblock


\bibitem[\protect\citeauthoryear{Taylor and Letham}{Taylor and Letham}{2017}]%
        {taylor2017forecasting}
\bibfield{author}{\bibinfo{person}{Sean~J Taylor} {and}
  \bibinfo{person}{Benjamin Letham}.} \bibinfo{year}{2017}\natexlab{}.
\newblock \showarticletitle{Forecasting at scale}.
\newblock \bibinfo{journal}{{\em The American Statistician\/}}
  (\bibinfo{year}{2017}).
\newblock


\bibitem[\protect\citeauthoryear{Uno, Asai, Uchida, and Arimura}{Uno
  et~al\mbox{.}}{2004}]%
        {Uno2004}
\bibfield{author}{\bibinfo{person}{Takeaki Uno}, \bibinfo{person}{Tatsuya
  Asai}, \bibinfo{person}{Yuzo Uchida}, {and} \bibinfo{person}{Hiroki
  Arimura}.} \bibinfo{year}{2004}\natexlab{}.
\newblock \bibinfo{booktitle}{{\em An Efficient Algorithm for Enumerating
  Closed Patterns in Transaction Databases}}.
\newblock


\bibitem[\protect\citeauthoryear{Zaki, Parthasarathy, Ogihara, Li,
  et~al\mbox{.}}{Zaki et~al\mbox{.}}{1997}]%
        {zaki1997new}
\bibfield{author}{\bibinfo{person}{Mohammed Zaki}, \bibinfo{person}{Srinivasan
  Parthasarathy}, \bibinfo{person}{Mitsunori Ogihara}, \bibinfo{person}{Wei
  Li}, {et~al\mbox{.}}} \bibinfo{year}{1997}\natexlab{}.
\newblock \showarticletitle{New Algorithms for Fast Discovery of Association
  Rules.}. In \bibinfo{booktitle}{{\em KDD}}.
\newblock


\bibitem[\protect\citeauthoryear{Zhang, Yuan, and Wang}{Zhang
  et~al\mbox{.}}{2014}]%
        {Zhang2014}
\bibfield{author}{\bibinfo{person}{Weinan Zhang}, \bibinfo{person}{Shuai Yuan},
  {and} \bibinfo{person}{Jun Wang}.} \bibinfo{year}{2014}\natexlab{}.
\newblock \showarticletitle{Optimal Real-time Bidding for Display Advertising}.
  In \bibinfo{booktitle}{{\em KDD'14}}.
\newblock


\bibitem[\protect\citeauthoryear{Zhang, Zhou, Wang, and Xu}{Zhang
  et~al\mbox{.}}{2016}]%
        {Zhang2016}
\bibfield{author}{\bibinfo{person}{Weinan Zhang}, \bibinfo{person}{Tianxiong
  Zhou}, \bibinfo{person}{Jun Wang}, {and} \bibinfo{person}{Jian Xu}.}
  \bibinfo{year}{2016}\natexlab{}.
\newblock \showarticletitle{Bid-aware Gradient Descent for Unbiased Learning
  with Censored Data in Display Advertising}. In \bibinfo{booktitle}{{\em KDD
  '16}}.
\newblock


\end{thebibliography}
